\documentclass{article}

\usepackage[final,nonatbib]{nips_2016}

\usepackage[utf8]{inputenc} % allow utf-8 input
\usepackage[T1]{fontenc}    % use 8-bit T1 fonts
\usepackage[pdftex,
            pdfusetitle,
            pdfauthor={John Glover},
            pdftitle={Modeling documents with Generative Adversarial Networks},
            pdfproducer={},
            pdfcreator={}]{hyperref}
\usepackage{url}            % simple URL typesetting
\usepackage{booktabs}       % professional-quality tables
\usepackage{amsfonts}       % blackboard math symbols
\usepackage{nicefrac}       % compact symbols for 1/2, etc.
\usepackage{microtype}      % microtypography
\usepackage{amsmath}
\usepackage[pdftex]{graphicx}
\usepackage[numbers]{natbib}
\setcitestyle{square}

\title{Modeling documents with Generative Adversarial Networks}

\author{
  John Glover \\
  Aylien Ltd. \\
  Dublin, Ireland \\
  \texttt{john@aylien.com} \\
}

\begin{document}

\maketitle

\begin{abstract}
This paper describes a method for using Generative Adversarial Networks to learn
distributed representations of natural language documents.
We propose a model that is based on the recently proposed Energy-Based GAN\@,
but instead uses a Denoising Autoencoder as the discriminator network.
Document representations are extracted from the hidden layer of the discriminator and
evaluated both quantitatively and qualitatively.
\end{abstract}

\section{Introduction}
\label{intro}

The ability to learn robust, reusable feature representations from unlabelled
data has potential applications in a wide variety of machine learning tasks,
such as data retrieval and classification.
One way to create such representations is to train deep generative models that can
learn to capture the complex distributions of real-world data.

In recent years, two effective approaches to training deep generative models have emerged.
The first is based on the Variational Autoencoder (VAE)~\cite{Kingma2013, Rezende2014},
where observed data \(x\) is assumed to be generated from a set of stochastic latent variables
\(z\).
The VAE introduces an inference network (implemented using a deep neural network) to approximate the intractable
distributions over \(z\), and then maximizes a lower bound on the log-likelihood of \(p(x)\).
The second approach is to use Generative Adversarial Networks (GANs)~\cite{Goodfellow2014a}.
In the original GAN formulation, a \textit{generator} deep neural network learns to map samples
from an arbitrary distribution to the observed data distribution.
A second deep neural network called the \textit{discriminator} is trained to distinguish between samples from the
empirical distribution and samples that are produced by the generator.
The generator is trained to create samples that will fool the discriminator, and so an adversarial game
is played between the two networks, converging on a saddle point that is a local minimum for the
discriminator and a local maximum for the generator.

Both VAE and GAN approaches have shown impressive results in producing generative models of
images~\cite{Radford2015,Gregor2015}, but relatively little work has been done on evaluating the performance of
these methods for learning representations of natural language.
Recently however, VAEs have been used successfully to create language models~\cite{Bowman2016}, to
model documents, and to perform question answering~\cite{Miao2016a}.
This paper investigates whether GANs can also be used to learn representations of natural language
in an unsupervised setting.
We describe a neural network architecture based on a variation of the recently proposed Energy-Based GAN
that is suitable for this task, provide a qualitative evaluation of the learned representations,
and quantitatively compare the performance of the model against a
strong baseline on a standard document retrieval task.

\subsection{Related work}
Representations for documents are often created by using generative topic models such as
Latent Dirichlet Allocation (LDA)~\cite{Blei2012}.
In LDA, documents consist of a mixture of topics, with each topic defining a
probability distribution over the words in the vocabulary. Each document is therefore
represented by a vector of mixture weights over its associated topics.

More recently, an undirected topic model based on the restricted Boltzmann machine (RBM)~\cite{Hinton2002}
called the Replicated Softmax~\cite{Salakhutdinov2009} was proposed.
Instead of viewing documents as distributions over topics, it forms a binary distributed
representation of each document, and was shown to outperform LDA both as a generative document
model and as a means of representing documents for a retrieval task.
One problem with the Replicated Softmax model is that it becomes too computationally expensive to
use it with larger vocabulary sizes, as the complexity scales linearly with the vocabulary size.
This was one of the factors that led to the development of an autoregressive neural topic model
called DocNADE~\cite{Larochelle}, which is based on the NADE model~\cite{Larochelle2011}.
The DocNADE model outperformed the Replicated Softmax when evaluated using the same document modelling
and retrieval benchmark, and had an added advantage in that the probability of an observation could be
computed exactly and efficiently. The most recent state-of-the-art for this task is a deeper
version of DocNADE~\cite{Lauly2016}.

\section{Learning representations with Generative Adversarial Networks}
The original GAN formulation~\cite{Goodfellow2014a} consists of a min-max adversarial game between a generative model \(G\)
and a discriminative model \(D\).
\(G(z)\) is a neural network, that is trained to map samples \(z\) from a prior noise distribution
\(p(z)\) to the data space.
\(D(x)\) is another neural network that takes a data sample \(x\) as input and outputs a single
scalar value representing the probability that \(x\) came from the data distribution instead of
\(G(z)\).
\(D\) is trained to maximise the probability of assigning the correct label to the input \(x\),
while \(G\) is trained to maximally confuse \(D\), using the gradient of \(D(x)\) with respect to
\(x\) to update its parameters.
The training procedure is described by Equation~\ref{eqn:gan}.
\begin{equation}
    \min_G \max_D E_{x \sim p(data)}[\log D(x)] + E_{z \sim p(z)}[\log (1 - D(G(z)))]
    \label{eqn:gan}
\end{equation}

One shortcoming with this model is that there is no explicit means for inference, and so it is
unclear how GANs should be used to do unsupervised representation learning.
In~\cite{Goodfellow2014a} two possible solutions are suggested, and have been explored by the
research community in subsequent works.
The first approach is to train another network to do inference,
learning a mapping from \(x\) back to \(z\)~\cite{Larsen2015,Dumoulin2016,Donahue2016},
with a variation on this method being to instead use the adversarial training process to regularize an
Autoencoder's representation layer~\cite{Shlens2016}.
The second idea is to use internal components of the discriminator network as a
representation~\cite{Radford2015}.
We investigated both approaches, but in initial experiments we found it difficult to find an
architecture that resulted in stable training across a range of datasets and model hyperparameters
when using a probabilistic discriminator network.
Performance improved significantly however when we switched to using the Energy-Based GAN
architecture, where the discriminator is an Autoencoder~\cite{Zhao2016}.
Document representations can then be formed from the encoded representation of the discriminator.
We describe this model concretely in Section~\ref{sec:adm}.

\section{An adversarial document model}
\label{sec:adm}

Let \(\mathbf{x} \in {\{0, 1\}}^{V}\) be the binary bag-of-words representation of a document, where
\(V\) is the vocabulary size and \(\mathbf{x}_i\) is a binary value indicating whether
the \(i\)\textsuperscript{th} word is present in the document or not.
We define a feedforward generator network \(G(\mathbf{z})\) that takes a vector \(\mathbf{z} \in \mathbb{R}^{h_g} \)
as input and produces a vector \(\mathbf{\hat{x}} \in \mathbb{R}^V \), with \(h_g\) being the number of dimensions
in the input noise vector (sampled from \(\mathcal{N}(0, I)\)).
We also define a discriminator network \(D(\mathbf{x})\), seen as an energy function, that takes vectors
\(\mathbf{x} \in \mathbb{R}^V \) and produces an energy estimate \(E \in \mathbb{R}\).

One difference to~\cite{Zhao2016} is that we use a Denoising Autoencoder (DAE) as our
energy function, as the DAE has been found to produce superior representations to the standard
Autoencoder~\cite{Vincent2010a}.
In this work we use single encoding and decoding layers, so the encoding process is
\begin{equation}
    \mathbf{h} = f(\mathbf{W}^e \mathbf{x}^c + \mathbf{b}_e)
    \label{eqn:enc}
\end{equation}
where \(\mathbf{W}^e\) is a set of learned parameters,
\(\mathbf{b}_e\) is a learned bias term,
\(f\) is a nonlinearity,
\(\mathbf{x}^c\) is a corrupted version of \(\mathbf{x}\),
and \(\mathbf{h} \in \mathbb{R}^{h_d}\) is the hidden representation of size \(h_d\).
The decoding process is given by
\begin{equation}
    \mathbf{y} = \mathbf{W}^d \mathbf{h} + \mathbf{b}_d
    \label{eqn:dec}
\end{equation}
where \(\mathbf{W}^d\) and \(\mathbf{b}_d\) are another learned set of weights and bias terms.
The final energy value is the mean squared reconstruction error:
\begin{equation}
    \frac{1}{V} \displaystyle\sum_{i=1}^{V} {(\mathbf{x}_i - \mathbf{y}_i)}^2
    \label{eqn:energy}
\end{equation}

\begin{figure}[h]
    \centering
    \includegraphics[width=0.8\linewidth]{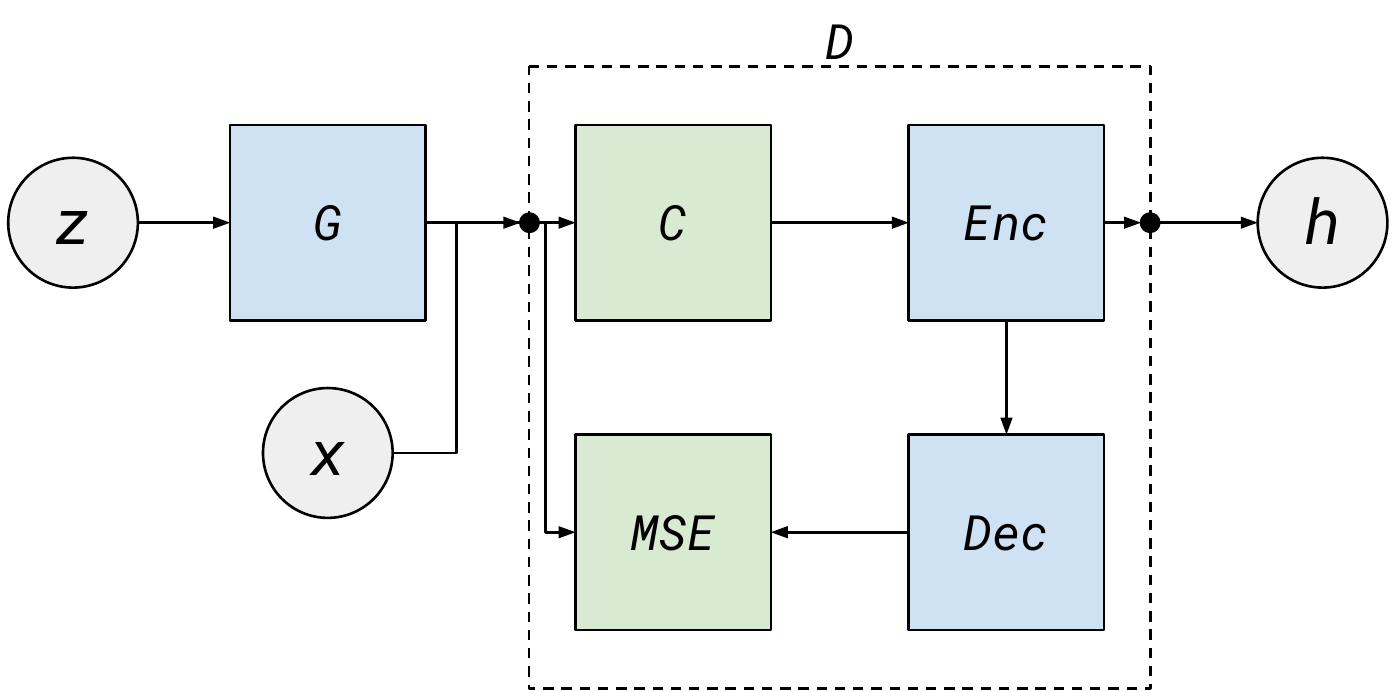}
    \caption{
        Using an Energy-Based GAN to learn document representations.
        \(G\) is the generator, \(Enc\) and \(Dec\) are
        DAE encoder and decoder networks,
        \(C\) is a corruption process (bypassed at test time) and
        \(D\) is the discriminator.
    }
\label{fig:adm}
\end{figure}

The model is depicted in Figure~\ref{fig:adm}.
The energy function is trained to push down on the energy of real samples \(\mathbf{x}\), and to push up on
the energy of generated samples \(\mathbf{\hat{x}}\)~\cite{Zhao2016}.
This is given by Equation~\ref{eqn:d_objective}, where \(f_D\) is the value to be minimised at each
iteration and \(m\) is a margin between positive and negative energies.
\begin{equation}
    f_{D}(\mathbf{x}, \mathbf{z}) = D(\mathbf{x}) + \max(0, m - D(G(\mathbf{z})))
    \label{eqn:d_objective}
\end{equation}
At each iteration, the generator \(G\) is trained adversarially against \(D\) to minimize \(f_G\):
\begin{equation}
    f_{G}(\mathbf{z}) = D(G(\mathbf{z}))
    \label{eqn:g_objective}
\end{equation}

\section{Experiments}
In this section we present experimental results based on the
20 Newsgroups\footnote{\url{http://qwone.com/~jason/20Newsgroups}} corpus, comparing the
adversarial document model to DocNADE\@.
20 Newsgroups consists of 18,786 documents (postings)
partitioned into 20 different newsgroups, where each document is assigned to a single topic.
The data is split into 11,314 training and 7,532 test documents.
We apply the standard preprocessing as in \citep{Salakhutdinov2009} and set the vocabulary size to 2000.

\subsection{Training details}
In order to make a direct comparison with \citep{Larochelle} we set our representation size \(h_d\)
(the size of the DAE hidden state) to 50. The generator input noise vector \(h_g\) is also set to be the same size.
The generator is a 3-layer feedforward network, with ReLU activations in the first 2 layers and a sigmoid nonlinearity
in the output layer. Layers 1 and 2 are both of size 300, with the final layer being the same size as
the vocabulary. Layers 1 and 2 use batch normalization~\cite{Ioffe2015}.
The discriminator encoder consists of a single linear layer followed by a leaky ReLU nonlinearity (with a leak of 0.02).
The decoder is a linear transformation back to the vocabularly size.
We optimize both \(G\) and \(D\) using Adam~\cite{Kingma2014a} with an initial learning rate of 0.0001.
Our DAE corruption process is to randomly zero 40\% of the input values, and we use a margin size \(m\) of
5\% of the vocabulary size.
We follow the same validation procedure as~\cite{Larochelle}, setting aside a validation set of 1000 documents
and using the average precision at 0.02\% retrieved documents as a performance measure for model selection.
DocNADE was trained using the publicly available code\footnote{\url{http://www.dmi.usherb.ca/~larocheh/code/docnade.zip}},
with a learning rate of 0.01 and using the tanh activation function.

\subsection{Document retrieval evaluation}
\begin{figure}[h]
    \centering
    \includegraphics[width=0.8\linewidth]{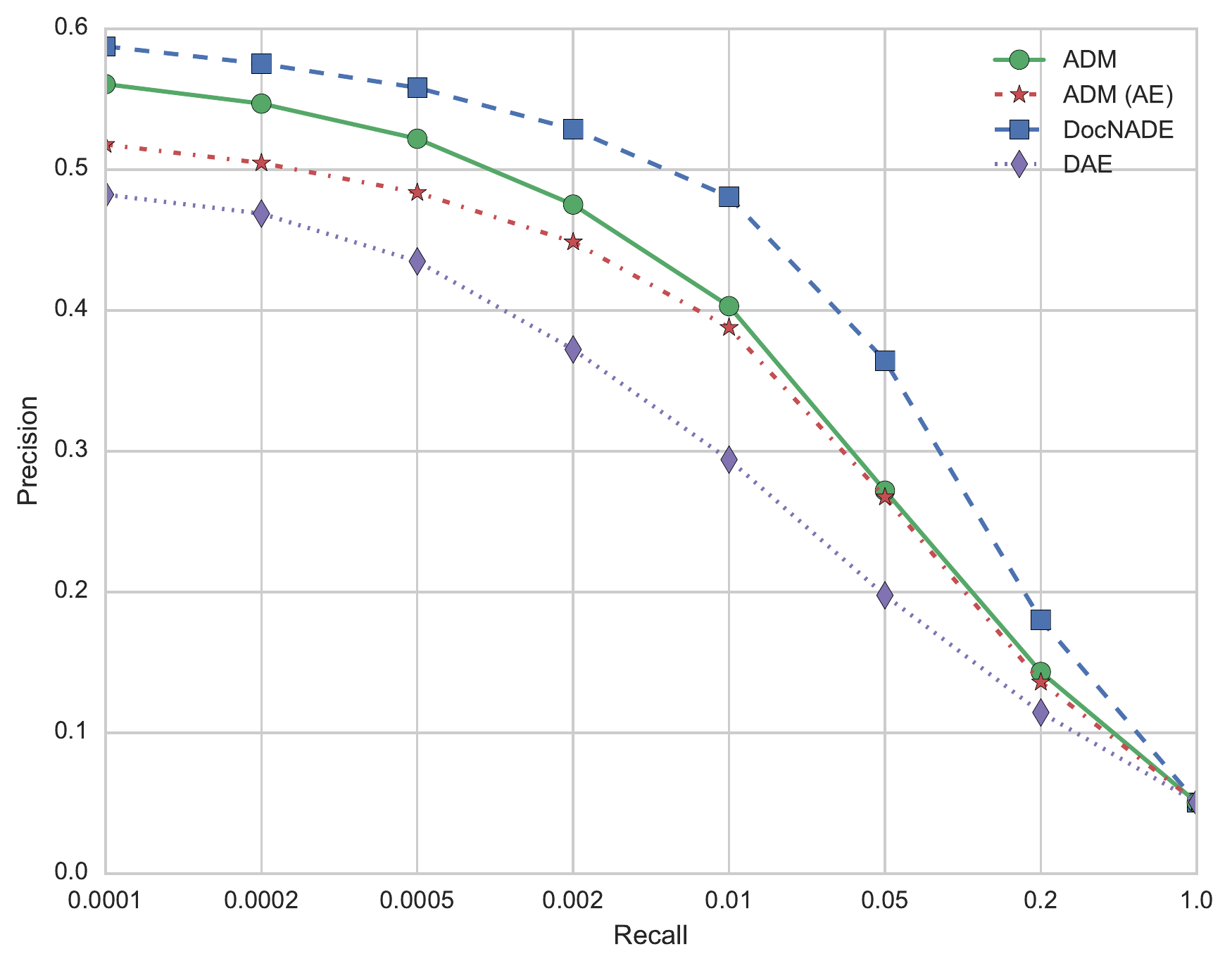}
    \caption{
        Precision-recall curves for the document retrieval task on the 20 Newsgroups dataset.
        ADM is the adversarial document model,
        ADM (AE) is the adversarial document model with a standard
        Autoencoder as the discriminator (and so it similar to the Energy-Based GAN~\cite{Zhao2016}),
        and DAE is a Denoising Autoencoder~\cite{Vincent2010a}.
    }
\label{fig:20ng_results}
\end{figure}

We performed the same document retrieval evaluation as in~\cite{Salakhutdinov2009,Larochelle,Lauly2016}.
All of the held-out test documents are used as queries and compared to a fraction of the closest documents in the
training set, where similarity is calculated using the cosine similarity between the
vector representations.
The average number of returned documents that have the same label as the query document (the precision) are recorded.
The results are shown in Figure~\ref{fig:20ng_results}.
In its current formulation, the adversarial document model still falls short of DocNADE's performance,
particularly for recall values between 0.002 and 0.05. It does approach DocNADE performance for lower
recall values however.
As the adversarial document model uses a DAE as a discriminator, for reference we also
include results obtained by just training a single layer DAE with the same corruption process, nonlinearity and mean-squared error loss, and also to a version of the adversarial document model that uses a standard Autoencoder instead of
a DAE as a discriminator.
Both of these models perform worse than the adversarial document model and DocNADE\@.

\subsection{Qualitative evaluation}
\begin{figure}[h]
    \centering
    \includegraphics[width=0.8\linewidth]{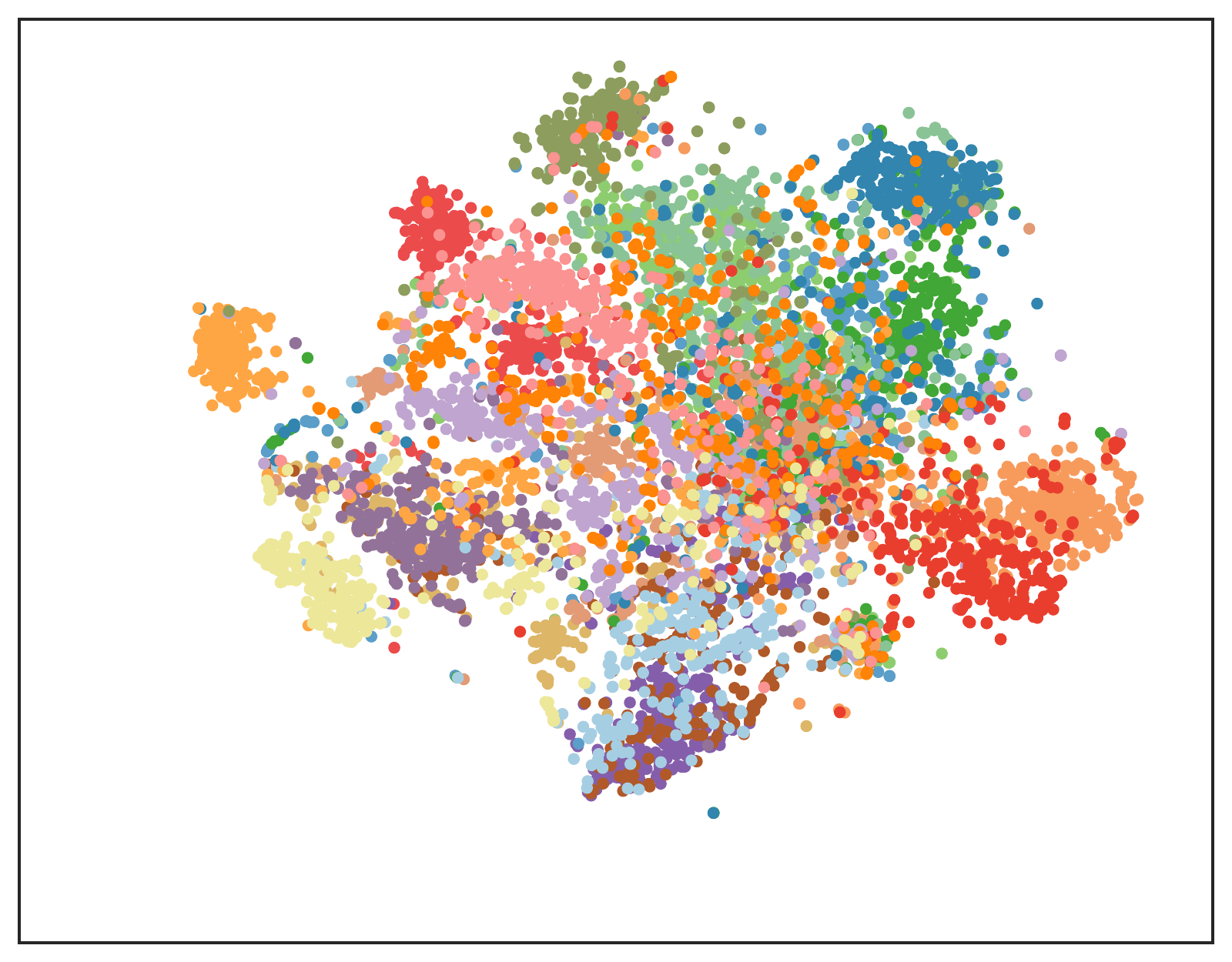}
    \caption{
        t-SNE visualizations of the document representations learned by the adversarial document model
        on the held-out test dataset of 20 Newsgroups.
        The documents belong to 20 different topics, which correspond to different coloured points in
        the figure.
    }
\label{fig:20ng_tsne}
\end{figure}

In this section we explore the semantic properties of the representations that are learned by the adversarial
document model.
For each hidden unit in the discriminator DAE, the weight values associated with each word in the vocabulary
can be viewed as the relative importance of a word to that particular hidden unit, and so the
hidden units can be interpreted as topics.
Table~\ref{tbl:topics} shows the words with the strongest absolute weight connections to selected hidden units
in the discriminator encoder layer, i.e.\ the words \(w\) that have the largest absolute values of
\(\mathbf{W}^{e}_{:,i}\) for a selected hidden unit \(i\).
We can deduce that these collections of words represent the topics of computing, sports and religion,
showing that the model is able to learn locally interpretable structure despite having no
interpretability constraints imposed on the representation layer.
However, in general we notice that the words that are strongly associated with each
hidden unit do not necessarily belong to a single coherent topic, as is evident by the inclusion of the words
\emph{bike} and \emph{rangers} in a collection of words related to religion.
Figure~\ref{fig:20ng_tsne} shows a visualization of the learned representations created using
t-SNE~\cite{VanDerMaaten2008}.

\begin{table}[t]
  \caption{20 Newsgroups hidden unit topics}
  \centering
  \begin{tabular}{lll}
    \toprule
    Computing & Sports & Religion \\
    \midrule
    windows & hockey & christians \\
    pc & season & windows \\
    modem & players & atheists \\
    scsi & baseball & waco \\
    quadra & rangers & batf \\
    floppy & braves & christ \\
    xlib & leafs & heart \\
    vga & sale & arguments \\
    xterm & handgun & bike \\
    shipping & bike & rangers \\
    \bottomrule
  \end{tabular}
\label{tbl:topics}
\end{table}

\section{Conclusion}
This paper shows that a variation on the recently proposed Energy-Based GAN can be used to learn
document representations in an unsupervised setting.
It also suggests some interesting areas for future research, including understanding why the
DAE in the GAN discriminator appears to produce significantly better representations than a
standalone DAE, and exploring the impact of applying additional constraints to the representation layer.

\bibliographystyle{ieeetr}
\bibliography{Bibliography}

\begin{thebibliography}{10}

\bibitem{Kingma2013}
D.~P. Kingma and M.~Welling, ``Auto-encoding variational bayes,'' {\em
  Proceedings of the International Conference on Learning Representations},
  2014.

\bibitem{Rezende2014}
D.~J. Rezende, S.~Mohamed, and D.~Wierstra, ``Stochastic backpropagation and
  approximate inference in deep generative models,'' {\em Proceedings of The
  31st International Conference on Machine Learning}, pp.~1278--1286, 2014.

\bibitem{Goodfellow2014a}
I.~J. Goodfellow, J.~Pouget-Abadie, M.~Mirza, B.~Xu, D.~Warde-Farley, S.~Ozair,
  A.~Courville, and Y.~Bengio, ``Generative adversarial networks,'' {\em
  Advances in Neural Information Processing Systems}, pp.~2672--2680, 2014.

\bibitem{Radford2015}
A.~Radford, L.~Metz, and S.~Chintala, ``Unsupervised representation learning
  with deep convolutional generative adversarial networks,'' {\em Proceedings
  of the International Conference on Learning Representations}, 2016.

\bibitem{Gregor2015}
K.~Gregor, I.~Danihelka, A.~Graves, D.~J. Rezende, and D.~Wierstra, ``Draw: A
  recurrent neural network for image generation,'' {\em Proceedings of The 32nd
  International Conference on Machine Learning}, 2015.

\bibitem{Bowman2016}
S.~R. Bowman, L.~Vilnis, O.~Vinyals, A.~M. Dai, R.~Jozefowicz, and S.~Bengio,
  ``Generating sentences from a continuous space,'' {\em Proceedings of the
  International Conference on Learning Representations}, 2016.

\bibitem{Miao2016a}
Y.~Miao, L.~Yu, and P.~Blunsom, ``Neural variational inference for text
  processing,'' {\em Proceedings of the International Conference on Learning
  Representations}, 2016.

\bibitem{Blei2012}
D.~M. Blei, A.~Y. Ng, and M.~I. Jordan, ``Latent dirichlet allocation,'' {\em
  Journal of Machine Learning Research}, vol.~3, no.~4-5, pp.~993--1022, 2012.

\bibitem{Hinton2002}
G.~E. Hinton, ``Training products of experts by minimizing contrastive
  divergence,'' {\em Neural computation}, vol.~14, no.~8, pp.~1771--1800, 2002.

\bibitem{Salakhutdinov2009}
R.~Salakhutdinov and G.~Hinton, ``Replicated softmax: An undirected topic
  model,'' {\em Advances in Neural Information Processing Systems},
  pp.~1607--1614, 2009.

\bibitem{Larochelle}
H.~Larochelle and S.~Lauly, ``A neural autoregressive topic model,'' {\em
  Advances in Neural Information Processing Systems}, pp.~2708--2716, 2012.

\bibitem{Larochelle2011}
H.~Larochelle and I.~Murray, ``The neural autoregressive distribution
  estimator,'' {\em International Conference on Machine Learning}, vol.~15,
  pp.~29--37, 2011.

\bibitem{Lauly2016}
S.~Lauly, Y.~Zheng, A.~Allauzen, and H.~Larochelle, ``Document neural
  autoregressive distribution estimation,'' {\em Journal of Machine Learning
  Research}, vol.~1, pp.~1--19, 2016.

\bibitem{Larsen2015}
A.~B.~L. Larsen, S.~K. S{\o}nderby, H.~Larochelle, and O.~Winther,
  ``Autoencoding beyond pixels using a learned similarity metric,'' {\em arXiv
  preprint arxiv:1512.09300}, 2015.

\bibitem{Dumoulin2016}
V.~Dumoulin, I.~Belghazi, B.~Poole, A.~Lamb, M.~Arjovsky, O.~Mastropietro, and
  A.~Courville, ``Adversarially learned inference,'' {\em arXiv preprint
  arxiv:1606.00704}, 2016.

\bibitem{Donahue2016}
J.~Donahue, P.~Kr{\"{a}}henb{\"{u}}hl, and T.~Darrell, ``Adversarial feature
  learning,'' {\em arXiv preprint arxiv:1605.09782}, 2016.

\bibitem{Shlens2016}
A.~Makhzani, J.~Shlens, N.~Jaitly, and I.~Goodfellow, ``Adversarial
  autoencoders,'' {\em International Conference on Learning Representations
  Workshop}, 2016.

\bibitem{Zhao2016}
J.~Zhao, M.~Mathieu, and Y.~LeCun, ``Energy-based generative adversarial
  network,'' {\em arXiv preprint arxiv:1609.03126}, 2016.

\bibitem{Vincent2010a}
P.~Vincent, H.~Larochelle, I.~Lajoie, Y.~Bengio, and P.-A. Manzagol, ``Stacked
  denoising autoencoders: Learning useful representations in a deep network
  with a local denoising criterion,'' {\em Journal of Machine Learning
  Research}, vol.~11, no.~3, pp.~3371--3408, 2010.

\bibitem{Ioffe2015}
S.~Ioffe and C.~Szegedy, ``Batch normalization: Accelerating deep network
  training by reducing internal covariate shift,'' {\em arXiv preprint
  arxiv:1502.03167}, 2015.

\bibitem{Kingma2014a}
D.~P. Kingma and J.~L. Ba, ``Adam: A method for stochastic optimization,'' {\em
  International Conference on Learning Representations}, 2015.

\bibitem{VanDerMaaten2008}
L.~{Van Der Maaten} and G.~Hinton, ``Visualizing high-dimensional data using
  t-sne,'' {\em Journal of Machine Learning Research}, vol.~9, pp.~2579--2605,
  2008.

\end{thebibliography}

\end{document}